\pdfoutput=1
\documentclass[11pt,a4paper]{article}
\usepackage[hyperref]{acl2020}
\usepackage{times}
\usepackage{latexsym}

\usepackage{url}
\usepackage[colorinlistoftodos,prependcaption,textsize=tiny]{todonotes}

\usepackage{amsmath}

\usepackage{amssymb}
\usepackage{subfig}

\aclfinalcopy 
\def\aclpaperid{1691} 

\title{Pre-training Is (Almost) All You Need: \\ An Application to Commonsense Reasoning}

\author{
  Alexandre Tamborrino\thanks{\hspace{0.15cm}Equal contribution.}\hspace{0.15cm}, Nicola Pellican\`{o}\footnotemark[1]\hspace{0.15cm}, Baptiste Pannier\footnotemark[1]\hspace{0.15cm}, \\ \textbf{Pascal Voitot, Louise Naudin} \\
  Samsung Strategy and Innovation Center \\
  \small{\texttt{\{a.tamborrino,n.pellicano,b.pannier,p.voitot,l.naudin\}@samsung.com}} \\}

\date{}

\begin{document}
\maketitle

\begin{abstract}
Fine-tuning of pre-trained transformer models has become the standard approach for solving common NLP tasks \cite{devlin2018bert}. Most of the existing approaches rely on a randomly initialized classifier on top of such networks. We argue that this fine-tuning procedure is sub-optimal as the pre-trained model has no prior on the specific classifier labels, while it might have already learned an intrinsic textual representation of the task. In this paper, we introduce a new scoring method that casts a plausibility ranking task in a full-text format and leverages the masked language modeling head tuned during the pre-training phase.
We study commonsense reasoning tasks where the model must rank a set of hypotheses given a premise, focusing on the COPA \cite{Gordon2011SemEval2012T7}, Swag \cite{zellers2018swagaf}, HellaSwag \cite{Zellers2019HellaSwagCA} and CommonsenseQA \cite{talmor2018commonsenseqa} datasets.
By exploiting our scoring method without fine-tuning, we are able to produce strong baselines (e.g. 80\% test accuracy on COPA) that are comparable to supervised approaches. Moreover, when fine-tuning directly on the proposed scoring function, we show that our method provides a much more stable training phase across random restarts (e.g $\times 10$ standard deviation reduction on COPA test accuracy) and requires less annotated data than the standard classifier approach to reach equivalent performances.
\end{abstract}

\section{Introduction}\label{sec:intro}
Recent advances in natural language processing have been made using sequential transfer learning over large pre-trained transformer models. From these models, most NLP tasks can be addressed by adding a classifier on top of the transformer embedding outputs \cite{devlin2018bert,Liu2019RoBERTaAR} .

In this paper, we tackle a subset of NLP tasks consisting in plausibility ranking. Such tasks can be formalised as follows: given a unique premise $p$ and a set of hypotheses  $H=\left\lbrace h_{i}\right\rbrace_{i=1 \dots n}$, the task consists in returning the appropriate hypothesis $h^{*} \in H$ that matches $p$ (see Section \ref{sec:met} for more details). A natural task that fits into this problem formulation is commonsense reasoning. Thus, it will be the main focus of the present paper.

\begin{figure*}[ht!]
  \centering
  \includegraphics[width=\linewidth]{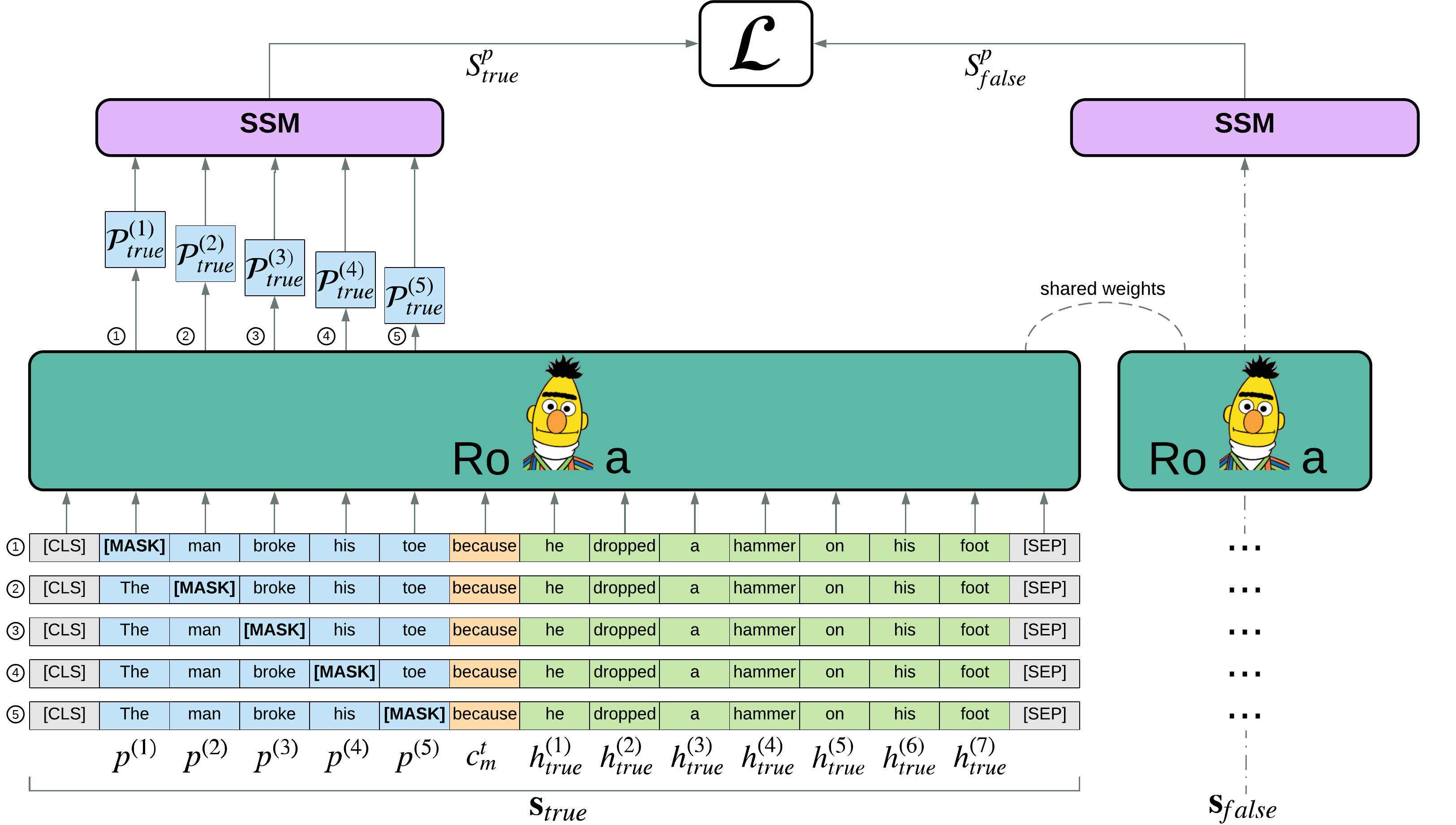}
  \caption{Overview of the proposed method for the task $t=$ COPA. Two \textit{full-text} sequences (Section \ref{sec:probform}), $\mathbf{s}_{true}$ and $\mathbf{s}_{false}$, are given as input (gold and distractor premise/hypothesis pairs respectively). Circled numbers explicitly mark input and output of five different versions of a given sentence, where each has a different premise word masked. The output probabilities $\mathcal{P}_{i}^{(k)}= P ( p^{(k)} \mid \mathbf{s}_{i}^{\backslash p^{(k)}})$ contribute to the score computation (\textit{target premise score} $S_{i}^{p}$ in this example, see Section \ref{sec:scoring}). When fine-tuning on the task is performed, gold and distractor scores are used for margin-based loss computation (Section \ref{sec:finetun}). }
  \label{fig:gros}
\end{figure*}

Traditionally, this problem is solved by jointly classifying each pair $(p, h_{i})_{i=1 \dots n}$. For instance, assuming a Masked Language Modeling (MLM) model is used, an example from the COPA dataset \cite{Gordon2011SemEval2012T7} is commonly casted into two distinct examples:
\begin{itemize}
\item \small{{\tt [CLS] The man broke his toe. [SEP] He dropped a hammer on his foot. [SEP]} $\rightarrow$ correct}
\item \small{{\tt [CLS] The man broke his toe. [SEP] He got a hole in his sock. [SEP]} $\rightarrow$ incorrect}
\end{itemize}
The special token {\tt [CLS]} (used for sentence level tasks) is then provided to a classifier in order to predict the label of the given example; {\tt [SEP]} is a special separator token. This format will be referred to as \textit{separated-sentence}.
For such a task, the use of the randomly initialized head can appear sub-optimal since the pre-trained model does not integrate any prior on the specific classifier label.
To validate this intuition, we cast the MLM model inputs into a \textit{full-text} format. Thus, the separation token is dropped and potentially replaced by conjunction words that are fully specific to the task. The previously illustrated correct example will be turned into: {\tt [CLS] The man broke his toe \underline{because} he dropped a hammer on his foot [SEP]}. Using this input format, we apply a new bidirectional word-level scoring function that leverages the MLM head \cite{devlin2018bert} tuned during the pre-training phase (see Figure \ref{fig:gros} for an overview of the proposed approach). This method produces strong zero-shot\footnote{For the following of our paper, we will note as \textit{zero-shot setting} the use of the pre-trained model without fine-tuning.} baselines on the COPA \cite{Gordon2011SemEval2012T7}, Swag \cite{zellers2018swagaf}, HellaSwag \cite{Zellers2019HellaSwagCA} and CommonsenseQA \cite{talmor2018commonsenseqa} datasets. Then, we fine-tune this new scoring function with a margin-based loss as proposed in \cite{li2019learning}. Using RoBERTa$_{LARGE}$, our results reveal that this new training procedure leads to better accuracy and much more stable training trajectories which is an important feature since large MLM models are known to be unstable on several tasks \cite{devlin2018bert, phang2018sentence}. Finally, we find that a progressive decrease of the training dataset size results in a progressive increase of the accuracy gap between our proposed method and the standard classifier ones. This makes our method advantageous in small dataset context.

\section{Related Work}\label{sec:rel}
In \cite{Trinh2018ASM}, researchers have shown that a RNN Language Model pretrained on a large amount of data can be used to efficiently score sentences in a zero-shot setting. They used the Winograd Schema Challenge (WSC-273) dataset \cite{Levesque2011TheWS} which mostly consists of a pronoun disambiguation task that requires commonsense reasoning. In their approach, the pronoun to disambiguate is replaced by the different candidates. Then, each version of the sentence is scored using the likelihood of the sequence under the forward autoregressive factorization. They showed that targeting the likelihood of the tokens placed after the candidate words performs better than a full-sentence likelihood estimation. This result highlights the fact that the choice of the targeted sub-sequence for the likelihood estimation has an important impact on the overall performance of the model. More recently, analysis of relational knowledge contained in pre-trained BERT models has been the subject of different studies \cite{petroni2019language, poerner2019bert}. Results have shown evidences that BERT models memorize reasoning about entity names and commonsense knowledge, making MLM models appropriate candidates to commonsense oriented tasks.

From a supervised learning perspective, \cite{li2019learning} proposed to replace the traditional cross-entropy loss with a margin-based one one the COPA dataset. The authors argued that cross-entropy based methods are not adapted for plausibility ranking tasks since they force the scores to adopt extreme values (near 0 or 1). In contrast, a  margin-based objective function appeared to be a natural way to rank a set of hypotheses. Both approaches were compared using the {\tt [CLS]} token of the BERT-base model and a \textit{separated-sentence} input format. The margin-based objective function surpassed the cross-entropy one by increasing the Test set accuracy from 73.4\% to 75.4\%.

Adopting a token level scoring approach \cite{Kocijan2019ASR} used a BERT model with a mixture between a margin-based and a MLM loss on WSC-273 to score the different pronouns to disambiguate. This approach allows the authors to improve the previous state of the art by 8.8\%. Despite being the closest method to the one proposed in this paper, our approach differs from three points:
\begin{itemize}
\item We generalize the scoring method by targeting different contiguous sub-sequences for the likelihood estimation. To do so, different datasets are recasted in a \textit{full-text format}.
\item We also focus on targeting the premise avoiding inner statistical biases of different hypotheses (e.g. word frequencies, punctuation, variable sequence lengths etc...).
\item The objective of the present paper is to propose a direct comparison in terms of accuracy and training stability across random restarts between the proposed method and standard classifers.
\end{itemize}

\section{Method}\label{sec:met}
\subsection{Problem Formulation} \label{sec:probform}
Given an input premise $p=(p^{(1)},p^{(2)},\dots,p^{(L_{p})}$), and a set of candidate hypotheses:
$$H=\left\lbrace h_{i} = (h_{i}^{(1)},h_{i}^{(2)}, \dots, h_{i}^{(L_{i})}) \right\rbrace_{i=1 \dots n},$$ we aim to identify the fitting hypothesis $h^{*} \in H$ which correctly matches $p$. The values $L_{p}$ and $\left \lbrace L_{i}  \right \rbrace_{i=1\dots n}$ are the sequence lengths of premise and hypotheses respectively. In a commonsense settings, such problem corresponds to find premise-hypothesis implications by exploiting some prior commonsense knowledge. Since our scoring method consumes input sequences in a \textit{full-text} format (see Section \ref{sec:scoring}), our method is formulated on a commonsense task but not limited to it.
\subsection{Sequence Scoring Method}\label{sec:scoring}
The proposed Sequence Scoring Method (SSM), takes as input a pair $   \langle p,h_{i} \rangle$ returns a score representing the likelihood of $h_{i}$ of being implied by $p$.

First, a transform operator $\mathcal{T}$ converts $\langle p,h_{i} \rangle$ pair into a \textit{full-text} input. Such operator, in it's simplest form, just concatenates the two sequences. However, in general $\mathcal{T}$ can be constrained on the task $t$.
\begin{equation}
\mathbf{s}_{i} = \mathcal{T}^{t}(p,h_{i}) = (c_{l}^t, p, c_{m}^t, h_{i}, c_{r}^t),
\end{equation}
where $\mathbf{s}_{i}$ is the resulting \textit{full-text} input, while $c_{l}^t$, $c_{m}^t$, and $c_{r}^t$ are left, middle and right conjunction sequences of the task. For example, Swag will have no conjunction, since the correct hypothesis is the natural continuation of the premise, while COPA will have \textit{because/so} middle conjunctions due to its cause/effect nature (see Section \ref{sec:copa}).

Given the \textit{full-text} input, the scorer aims to exploit the pre-training task of word masking in order to compute its result. Let us consider the masking of a word $w$ which contributes to make sense of the matching between $p$ and $h_{i}$. The intuition is that the confidence of the network in recovering such word is directly related to the score of $\langle p,h_{i} \rangle$. Let us define, inspired by the notation of \cite{song2019mass}, $\mathbf{s}_{i}^{\backslash w}$ as the sentence $\mathbf{s}_{i}$ with the tokens of $w$ replaced by the {\tt [MASK]} token.

The \textit{target premise score} is calculated as follows:
\begin{equation}\label{eq:tps}
    S^{p}_{i} = \sum_{k=1}^{L_{p}} \log \left[ P \left( p^{(k)} \mid \mathbf{s}_{i}^{\backslash p^{(k)}}\right)\right],
\end{equation}
where premise words are masked one by one in order to compute their relevance with respect to the given hypothesis. Masked word probability is estimated from direct inference on a model pre-trained on MLM task. The computational complexity of such method grows linearly with $L_{p}$ (requiring $L_{p}$ examples per forward pass). Alternatively, the \textit{target hypothesis score} is computed as:
\begin{equation}
    S^{h}_{i} = \frac{1}{L_{i}} \sum_{k=1}^{L_{i}} \log \left[ P \left( h_{i}^{(k)} \mid \mathbf{s}_{i}^{\backslash h_{i}^{(k)}}\right)\right].
\end{equation}
The \textit{target hypothesis score} needs normalization by $L_{i}$ in order to allow comparison between variable candidate hypothesis length.
The best hypothesis will be taken as the one maximizing the target premise (or hypothesis) score:
\begin{equation}
    h^{*} = h_{j} \in H \;\; s.t. \;\; \max_{i=1\dots n} S^{p}_{i} = S^{p}_{j}.
\end{equation}
As demonstrated in Section~\ref{sec:zero}, the \textit{target premise score} allows for a fairer comparison between different hypotheses. In fact, they present inherent differences in terms of statistical frequency of words, sequence length or may exhibit more or less strong inter-dependency between words (e.g. composite words reinforce each other confidence). Such variance could introduce a bias in the relative significance of each hypothesis alone (independently from the premise). On the opposite, different probabilities on the same target premise word can only be affected by the change of hypothesis context.

\subsubsection*{N-grams sequence scoring}
We can extend the proposed SSM by scoring the reconstruction not only of single words, but of entire n-grams. Adding n-grams probabilities to the logarithmic mean combination not only robustifies the scoring methods, but helps to better model the joint probability of (dependent) close words, especially in a zero-shot setting. Let us note as $p^{(u:v)}$ as the sub-sequence of $p$ spanning between indexes $u$ and $v$ (included).
The \textit{partial target premise score} for g-grams (i.e. mask windows of size $g$) can be expressed as:
$$
    S^{p,g}_{i} = \sum_{k=1}^{L_{p}-g+1} \log \left[ P \left( p^{(k:k+g-1)} \mid \mathbf{s}_{i}^{\backslash p^{(k:k+g-1)}}\right)\right].
$$
By definition the target premise score in Equation~\ref{eq:tps} is equivalent to 1-gram partial target premise score (i.e. $S^{p}_{i} \triangleq S^{p,1}_{i}$).
The n-gram sequence scoring accumulates masked language model probabilities from every gram size till $n$.
\begin{equation}
    S^{p,[1,n]}_{i} = \sum_{g=1}^{n}  S^{p,g}_{i}.
\end{equation}

\subsection{SSM-based fine-tuning}\label{sec:finetun}
The proposed score function, since it does not imply any addition of a head module, can be directly applied without any retraining (see Section~\ref{sec:zero}). It can also be directly used when fine-tuning on the task. The different masked inputs needed to compute the \textit{target premise score}, $\left \lbrace \mathbf{s}_{i}^{\backslash p^{(j)}}  \right \rbrace_{j=1..L_{p}}$, are batched together in order to compute score $S^{p}_{i}$ in one forward pass. The model acts as a siamese network that performs independent computation of \textit{target premise score} for each hypothesis $h_{i}$.

\subsubsection*{Loss function}\label{sec:loss}
As already noted in \cite{li2019learning}, multiple choice tasks (e.g. COPA) are more naturally expressed as \textit{learning to rank} problems. For this reason we adopt as objective function a \textbf{margin-based loss} in contrast to cross-entropy loss. Given ground truth sentence index $i^{*}$, the loss is specified as:
\begin{equation}
    \mathcal{L} = \frac{1}{n} \sum_{\substack{i=1 \\ i\neq i^{*}}}^{n} \max\left( 0, \eta - S^{p}_{i^{*}} + S^{p}_{i} \right),
\end{equation}
where $\eta$ is a margin threshold hyperparameter.

According to our preliminary experiments, we do not add a second MLM component in the general loss (as in \cite{Kocijan2019ASR}), since it always leads to a decrease of the model performance for various weighted contributions of the MLM term.

\section{Datasets}\label{sec:datasets}
The commonsense reasoning datasets that we focus on are COPA \cite{Gordon2011SemEval2012T7}, Swag \cite{zellers2018swagaf}, HellaSwag \cite{Zellers2019HellaSwagCA} and CommonsenseQA \cite{talmor2018commonsenseqa}. All these datasets share the premise-hypothesis task format. Table~\ref{tab:datasets_zs} shows examples of \textit{full-text} format and \textit{separated-sentence} format for all datasets.

\begin{table*}[t!]
\centering
\begin{tabular}{lp{60mm}p{60mm}}
  \textbf{Dataset} & \textbf{Full-text format} & \textbf{Separated-sentence format} \\
  \hline
  COPA (effect) & \small{{\tt [CLS]} I knocked on my neighbor's door \textbf{\textcolor{red}{so}} my neighbor invited me in. {\tt [SEP]}}& \small{{\tt [CLS]} I knocked on my neighbor's door. {\tt [SEP]} My neighbor invited me in. {\tt [SEP]}} \\ \hline
  COPA (cause) & \small{{\tt [CLS]} The man broke his toe \textbf{\textcolor{red}{because}} he dropped a hammer on his foot. {\tt [SEP]}} &  \small{{\tt [CLS]} He dropped a hammer on his foot. {\tt [SEP]} The man broke his toe. {\tt [SEP]}} \\ \hline
  CommonsenseQA & \small{{\tt [CLS]} \textit{\textcolor{blue}{Q:}} Where on a river can you hold a cup upright to catch water on a sunny day? \textbf{\textcolor{red}{A:}} waterfall {\tt [SEP]}} & \small{{\tt [CLS]} Q: Where on a river can you hold a cup upright to catch water on a sunny day? {\tt [SEP]} A: waterfall {\tt [SEP]}} \\ \hline
  Swag & \small{{\tt [CLS]} We notice a man in a kayak and a yellow helmet coming in from the left. As he approaches, his kayak flips upside-down. {\tt [SEP]}}& \small{{\tt [CLS]} We notice a man in a kayak and a yellow helmet coming in from the left. {\tt [SEP]} As he approaches, his kayak flips upside-down. {\tt [SEP]}} \\ \hline
  HellaSwag & \small{{\tt [CLS]} A man is standing in front of a camera. He starts playing a harmonica for the camera. He rocks back and forth to the music as he goes. {\tt [SEP]}}& \small{{\tt [CLS]} A man is standing in front of a camera. He starts playing a harmonica for the camera. {\tt [SEP]} He rocks back and forth to the music as he goes. {\tt [SEP]}}
\end{tabular}
\caption{Examples of \textit{full-text} format and \textit{separated-sentence} format for gold premise-hypothesis pairs. Left conjunction $c_{l}^t$ is highlighted in italic blue, middle conjunction $c_{m}^t$ in bold red.}
\label{tab:datasets_zs}
\end{table*}

\subsubsection*{COPA}\label{sec:copa}
COPA (Choice of Plausible Alternatives) \cite{Gordon2011SemEval2012T7} is a commonsense causal reasoning task where two candidate hypotheses are given. COPA itself is composed of two sub-tasks: \textit{effect} samples and \textit{cause} samples. The \textit{effect} and \textit{cause} samples have respectively \textit{implies} and \textit{implied by} relation with the correct hypothesis. The \textit{full-text} format of COPA is built by using the conjunction words {\tt because} (resp. {\tt so}) as middle conjunctions for \textit{cause} (resp. \textit{effect}) samples. Concerning the \textit{separated-sentence} format, we reverse the premise and hypothesis order for \textit{cause} samples in order to convert all \textit{cause} samples into \textit{effect} samples. This has the benefit to present a unique task to the model, and our experiments show that this give better results than keeping \textit{cause} samples and \textit{effect} samples unmodified. We choose the SuperGLUE split \cite{Wang2019SuperGLUEAS}.

\subsubsection*{CommonsenseQA}\label{sec:csqa}
CommonsenseQA \cite{talmor2018commonsenseqa} is a multiple-choice commonsense question answering dataset where each question has one correct answer and four distractor answers. To create the \textit{full-text} format, we prepend {\tt Q:\textvisiblespace} to the question, {\tt A:\textvisiblespace} to the answer, and then concatenate the question and the answer ($\textvisiblespace$ stands for space character). For the \textit{separated-sentence} format, we also use the {\tt Q:\textvisiblespace} and {\tt A:\textvisiblespace} prefixes to follow the best recommendation from the FairSeq repo on how to fine-tune RoBERTa on CommonsenseQA \footnote{https://github.com/pytorch/fairseq/tree/\\master/examples/roberta/commonsense\_qa}. Since the benchmark Test set is private, for our zero-shot and fine-tuning stability studies we have split the original validation set evenly, treating last 611 samples as Test set Test$^*$.

\subsubsection*{Swag and HellaSwag}\label{sec:swag}
Swag (Situations With Adversarial Generations) \cite{zellers2018swagaf} is a multiple choice commonsense dataset about grounded situations. Each premise is a video caption with four answer choices about what might happen next in the scene. The correct answer is the video caption for the next event in the video. The other negative answers are created via Adversarial Filtering: generated by language modeling models and filtered by discriminator models. HellaSwag \cite{Zellers2019HellaSwagCA} is an evolved version of Swag using better generators and discriminators models for Adversarial Filtering. Since the benchmark test set is private, we evaluate our zero-shot setting on the Val set (we do not perform a fine-tuning study on Swag and HellaSwag as explained in Section~\ref{sec:finetuning_results}).

\section{Experiments}\label{sec:exp}
In this section we first apply our scoring method in a zero-shot setting on the four aforementioned datasets. Then we fine-tune our scoring method while varying the percentage of the training data used and compare it to approaches that use a randomly initialized classifier head. We use RoBERTa$_{LARGE}$ \cite{Liu2019RoBERTaAR} for our pre-trained model as RoBERTa$_{LARGE}$ fine-tuned with a classification layer on top has very competitive results on those datasets. Our implementation use PyTorch and the HuggingFace Transformers library \cite{Wolf2019HuggingFacesTS}.

\subsection{Task probing}\label{sec:probing}
Before assessing our zero-shot and fine-tuning results, we perform a task probing by evaluating the zero-shot score we obtain by removing the premise from the input and only scoring the hypotheses. If the score is significantly better than a random baseline, it means that the task is not actually solved by commonsense reasoning, but by using statistical biases in the hypotheses. This probing method has been already used on several datasets to show that the underlying task was not really solved by the top-performing models \cite{Niven2019ProbingNN, Zellers2019HellaSwagCA}.

The results of the task probing evaluation are reported in Table~\ref{tab:probing}. While COPA and CommonsenseQA have a \textit{hypothesis only} score close to the random baseline, the score of both Swag and HellaSwag are significantly higher than their random baseline (more than twice). This confirms the study from \cite{Zellers2019HellaSwagCA} that shows that Swag's false hypotheses were generated using a weak generator, therefore the authors argue that the fine-tuning process on a BERT model on Swag learns to pick up the statistical cues left by the weak generator. Our results show that RoBERTa$_{LARGE}$ can leverage these distributional biases without the fine-tuning phase. We argue that the human-written pre-training corpora of RoBERTa biases it to give better score to human-written language rather than model-generated sentences. As shown in \cite{Holtzman2019TheCC}, there is indeed still a strong distributional differences between human text and machine text. Furthermore, our result also highlights that HellaSwag still exhibits a strong bias due to its generation scheme when evaluated with RoBERTa$_{LARGE}$.

\begin{table}[t!]
\begin{tabular}{lll}
    \textbf{Dataset} & \textbf{Mode} & \textbf{Acc$^1$ (\%)} \\
    \hline
    COPA & hyp-only & 54.6 \\
         & random & 50.0 \\
    \hline
    CommonsenseQA & hyp-only & 22.0 \\
                  & random & 20.0 \\
    \hline
    Swag & hyp-only & 60.6 \\
         & random & 25.0 \\
    \hline
    HellaSwag & hyp-only & 50.8 \\
              & random & 25.0 \\
\end{tabular}
\caption{Commonsense reasoning task probing. \textit{hyp-only} stands for hypothesis only, \textit{random} for random baseline. $^{1}$COPA is evaluated on Test, CommonsenseQA is evaluated on Test$^{*}$, Swag and HellaSwag are evaluated on Val (see Section~\ref{sec:csqa}).}
\label{tab:probing}
\end{table}

\subsection{Zero-shot Results}\label{sec:zero}
For both COPA and CommonsenseQA, the best performing scoring method uses the \textit{target premise} and 4-grams settings as shown in Tables~\ref{tab:copa_zs} and \ref{tab:cmqa_zs}. Targeting the premise gives better results than targeting the hypothesis, which reinforces our argument that targeting the hypothesis may be harder as the differences between the hypotheses make the score comparison noisier. Also, more grams give increasingly better results but the trend inverts after 4-grams, which may be due to the fact that masked models are not trained to mask large chunks of text.
It is interesting to note that our zero-shot result is significantly better than a BERT$_{LARGE}$ cross-entropy model fined-tuned on the COPA training set (80.0\% vs. 70.6\% accuracy) \cite{Wang2019SuperGLUEAS}, while being comparable for CommonsenseQA \footnote{https://www.tau-nlp.org/csqa-leaderboard}. Moreover, when we intentionally switch the {\tt so} and {\tt because} conjunction words on COPA to make the samples erroneous, the accuracy drops significantly (64.4\%). We reckon this is an indicator that our scoring method effectively reuse the pre-learned representation the \textit{full-text} format of the task.

\begin{table}[t!]
\begin{tabular}{lll}
    \textbf{Target} & \textbf{Grams} & \textbf{Test Acc (\%)} \\ \hline
    premise & 1 & 74.0 \\
    hypothesis & 1 & 69.8 \\
    premise & 2 & 76.2 \\
    premise & 3 & 79.0 \\
    premise & 4 & \textbf{80.0} \\
    premise & 5 & 79.4 \\
\end{tabular}
\caption{COPA zero-shot results.}
\label{tab:copa_zs}
\end{table}

\begin{table}[t!]
\begin{tabular}{lll}
    \textbf{Target} & \textbf{Grams} & \textbf{Test$^{*}$ Acc (\%)} \\ \hline
    premise & 1 & 47.8 \\
    hypothesis & 1 & 37.4 \\
    premise & 2 & 53.2 \\
    premise & 3 & 53.7 \\
    premise & 4 & \textbf{56.1} \\
    premise & 5 & 55.2 \\
\end{tabular}
\caption{CommonsenseQA zero-shot results.}
\label{tab:cmqa_zs}
\end{table}

Concerning Swag and HellaSwag, the \textit{target hypothesis} mode is significantly better than the \textit{target premise} mode (see Table~\ref{tab:swag_zs}), as expected from our task probing work in Section~\ref{sec:probing}. For example, on HellaSwag, the \textit{target hypothesis} mode is only 8\% better than the \textit{hypothesis only} mode (58.8\% versus 50.8\%), which confirms that on this setting our zero-shot method is mainly taking advantage of the bias in the hypotheses. Therefore we refrain from doing more zero-shot experiments on both datasets.

\begin{table}[t!]
\begin{tabular}{lll}
    \textbf{Dataset} & \textbf{Target} & \textbf{Val Acc (\%)} \\ \hline
    Swag & premise & 48.3 \\
    Swag & hypothesis & \textbf{72.5} \\ \hline
    HellaSwag & premise & 37.1 \\
    HellaSwag & hypothesis & \textbf{58.8} \\
\end{tabular}
\caption{Swag/HellaSwag zero-shot results (1-Gram).}
\label{tab:swag_zs}
\end{table}

\subsection{Fine-tuning Results}\label{sec:finetuning_results}

\begin{figure*} [ht!]
\captionsetup[subfloat]{farskip=0pt,captionskip=0.1pt}
\centering
\subfloat[]{\includegraphics[height=6.2cm]{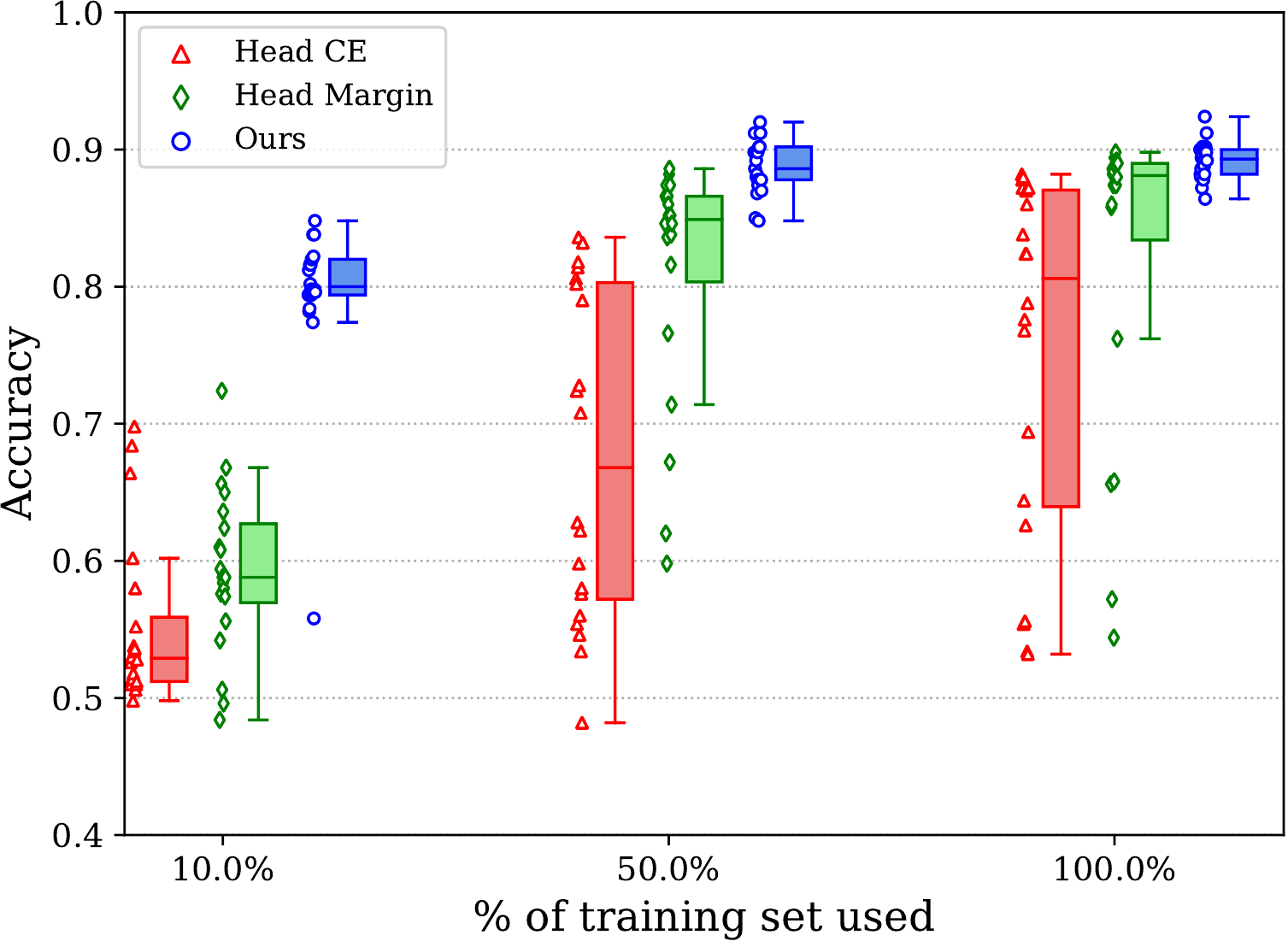}\label{fig:copa_finetuning_a}}
\subfloat[]{\includegraphics[height=6.2cm]{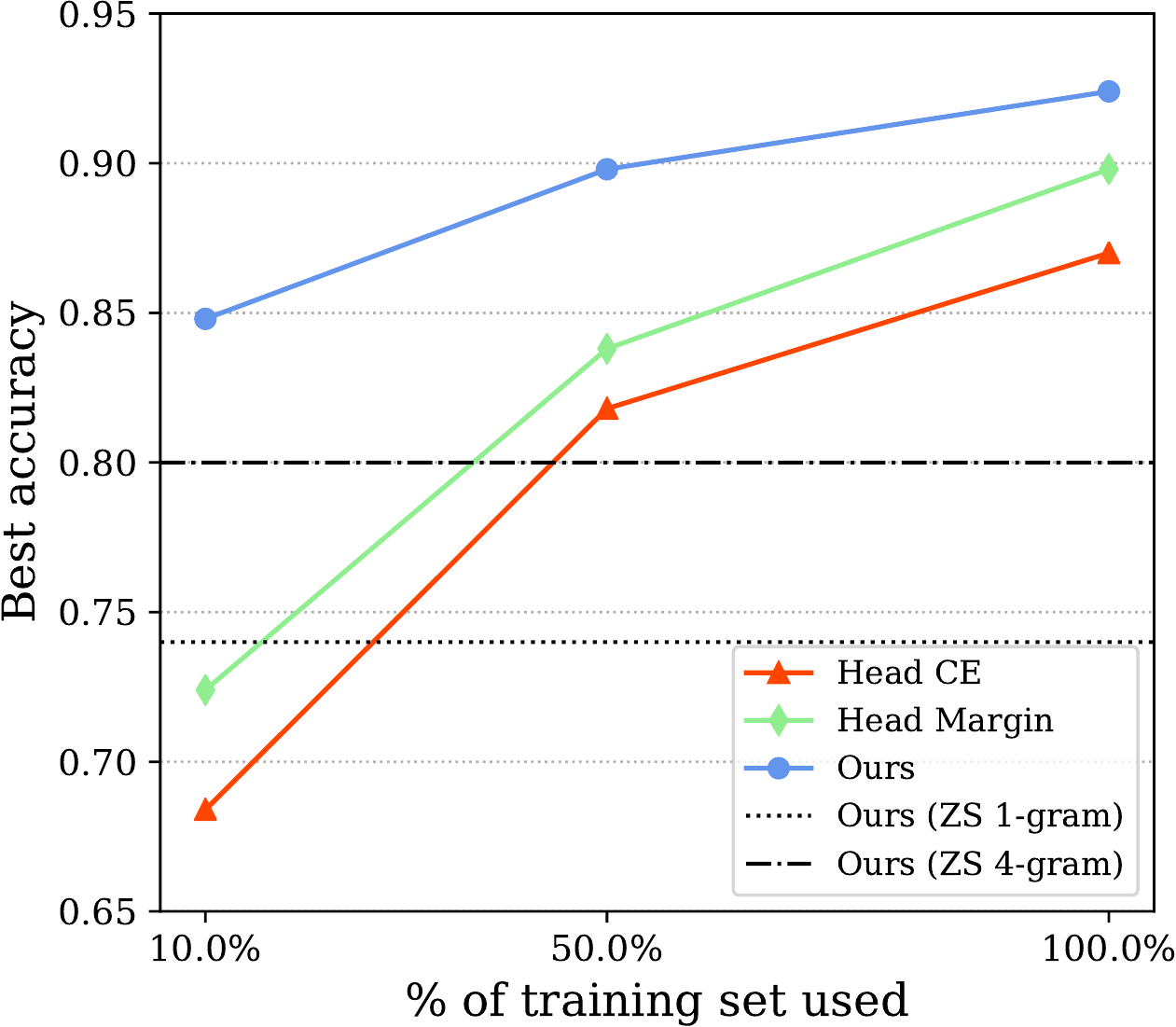}\label{fig:copa_finetuning_b}}
\caption{COPA fine-tuning results on Test set. The whole training set corresponds to 400 examples.}
\label{fig:copa_finetuning}
\end{figure*}

\begin{figure*} [ht!]
\captionsetup[subfloat]{farskip=0pt,captionskip=0.1pt}
\centering
\subfloat[]{\includegraphics[height=6.2cm]{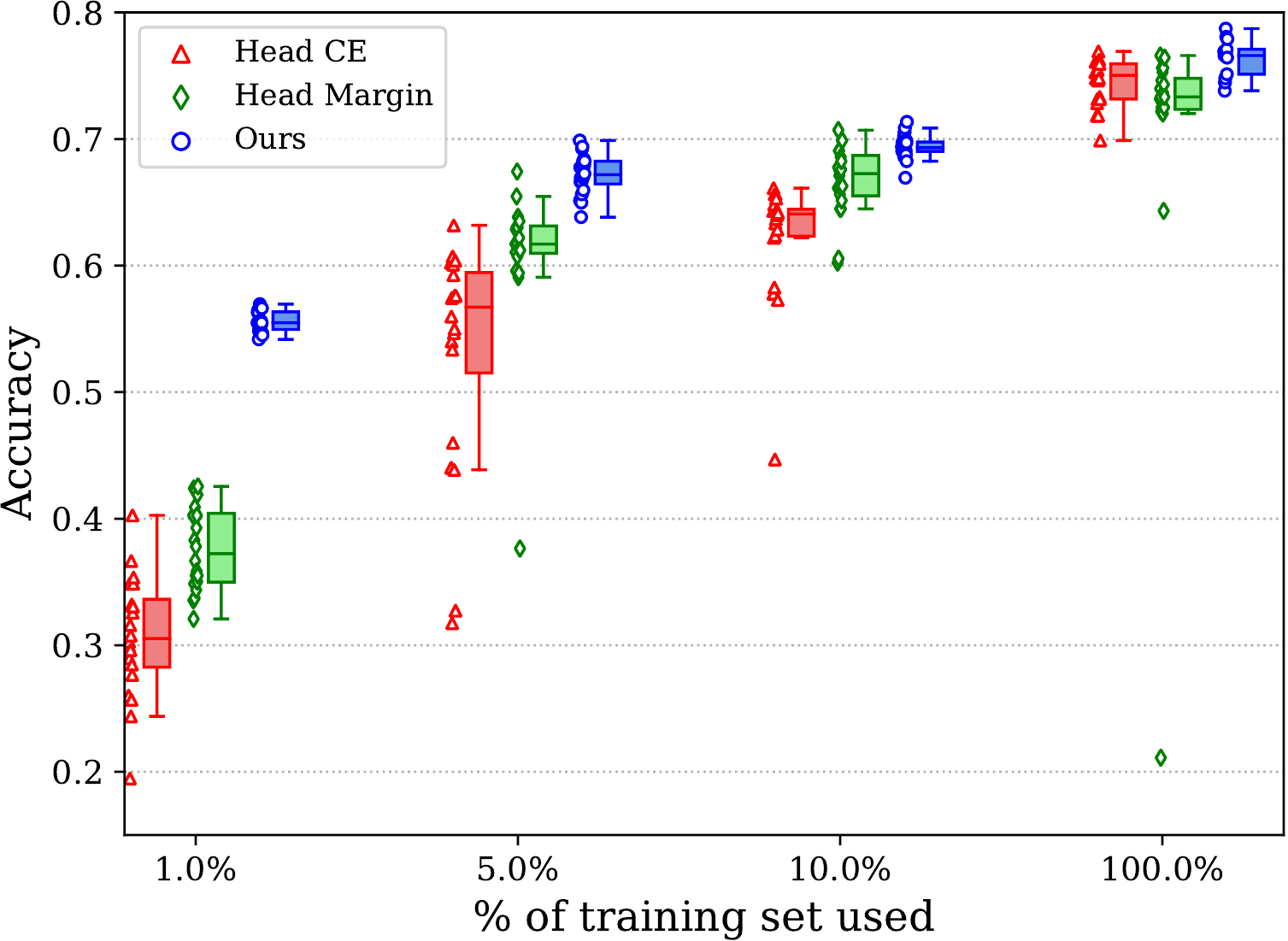}\label{fig:cmqa_finetuning_a}}
\subfloat[]{\includegraphics[height=6.2cm]{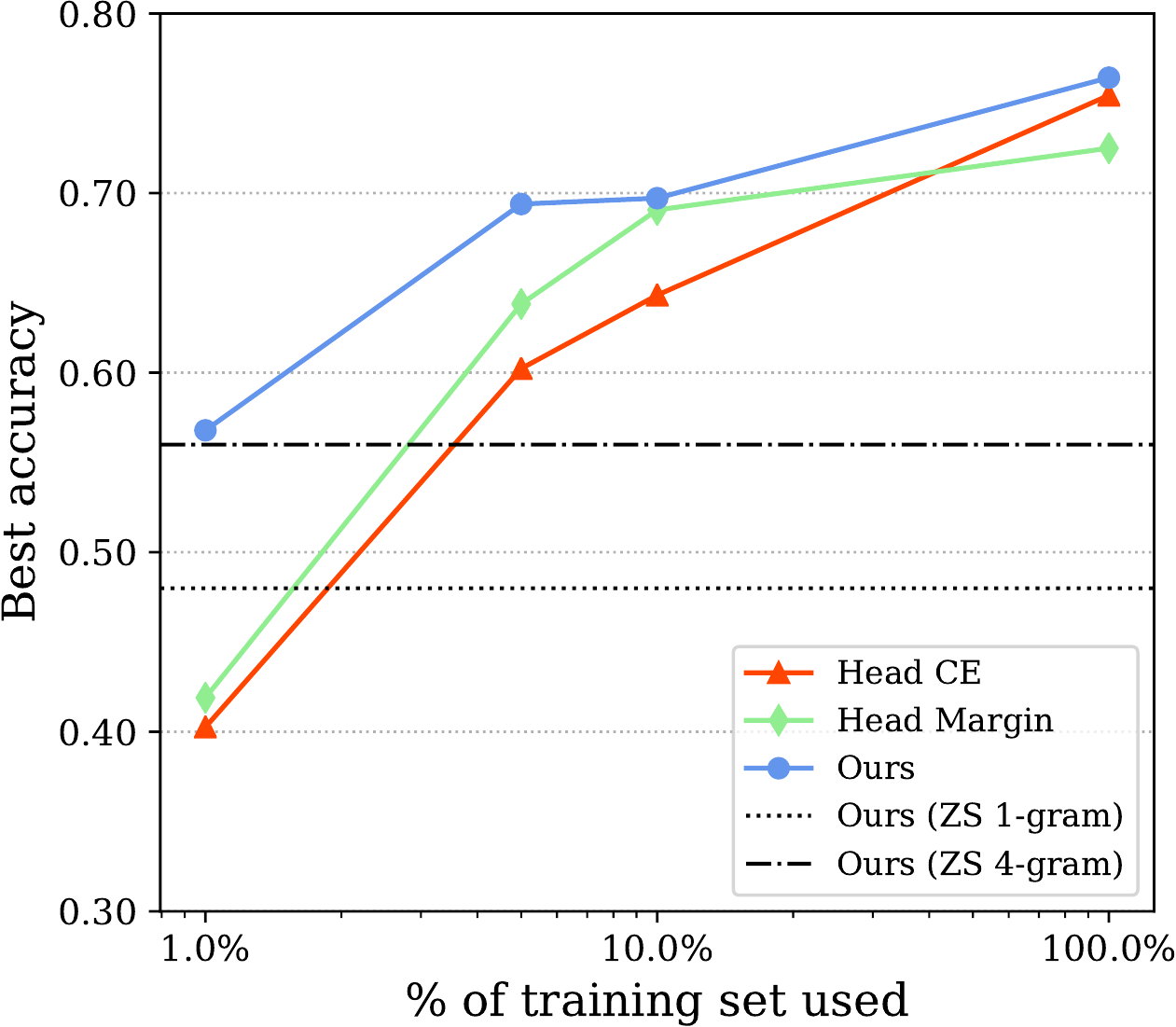}\label{fig:cmqa_finetuning_b}}
\caption{CommonsenseQA fine-tuning results on Test$^*$ set. The whole training set corresponds to 9741 examples.}
\label{fig:cmqa_finetuning}
\end{figure*}

Following the strong bias of Swag and HellaSwag that was shown in Section~\ref{sec:probing} using our scoring method with RoBERTa$_{LARGE}$, we decide to not include them into our fine-tuning study to be sure to compare results for which models learn the actual premise-hypothesis commonsense reasoning task.

\subsubsection*{Comparison settings}

In order to make fair comparisons, we train and compare three different model settings:
\begin{itemize}
  \item Our scoring method with \textit{target premise} mode, 1-gram, margin-based loss, \textit{full-text} format (\textit{ours}).
  \item A randomly initialized classifier with cross-entropy loss and \textit{separated-sentence} format (\textit{head CE}). The cross-entropy loss is computed on the probability of the correct candidate, normalized over all candidates in the set (see Equation 1 in \cite{li2019learning}).
  \item A randomly initialized classifier with margin-based loss and \textit{full-text} format (\textit{head margin})
\end{itemize}

The \textit{head margin} setting is an ablated version of our scoring method to verify that our reuse of the MLM head actually provides a significant advantage over a randomly initialized head.
For our method, we report results only for the best performing scoring method which is the \textit{target premise} mode. Experiments showed us that varying the number of grams produce comparable results, so we use the 1-gram setting for computational efficiency. We reckon that the enriched bi-directional context granted by N-gram score can be directly learned when fine-tuning on the task.

For each dataset, we train the three model settings for 20 random seeds each. For each seed, we pick the best performing model on the validation set and report its accuracy on the Test set. We then compute the max accuracy, mean accuracy and standard deviation of each model setting on the Test set.
For all model settings, following the recommended hyper-parameters to fine-tune RoBERTa$_{LARGE}$ \cite{Liu2019RoBERTaAR}, we set a learning rate of 1e-5, a warm-up ratio of 6\% of the total number of training steps, a linear learning rate decay and a weight decay of 0.01. We use a batch size of 8 for COPA (4 for the 10\% training percentage setting) and 16 for CommonsenseQA. For the margin-based loss (\textit{ours} and \textit{head margin}), we set $\eta = 0.5$ after a few trials.

\subsubsection*{COPA and CommonsenseQA results}

On both COPA and CommonsenseQA, our method outperforms both the \textit{head CE} and \textit{head margin} methods in terms of mean accuracy and max/best accuracy (see Figure~\ref{fig:copa_finetuning} and Figure~\ref{fig:cmqa_finetuning}). Moreover, we find that a progressive decrease of the training dataset size results in a progressive increase of the best accuracy gap between our method and the other ones. This confirms our intuition that our methods is the most advantageous when few training data is available.

For example, when using 1\% of training data of CommonsenseQA, our method achieves an accuracy of 56.7\% on the Test$^{*}$ set (vs. 40.2\% for the \textit{head CE} approach). Using the whole training data, our approach still outperforms other methods but by a lower margin (76.4\% accuracy versus 75.4\% for \textit{head CE}). In addition, when evaluated on the CommonsenseQA private Test set, our approach gets 71.6\% accuracy which is close to RoBERTa$_{LARGE}$ cross-entropy \cite{Liu2019RoBERTaAR} under an important hyper-parameter grid search\footnote{https://github.com/pytorch/fairseq/tree/\\master/examples/roberta/commonsense\_qa} (72.1\% accuracy).

When using 100\% of the COPA training set (400 train samples), our method outperforms the \textit{head CE} setting per 5 points and the \textit{head margin} setting per 3 points, achieving an accuracy of 92.4\% on the Test set. This result allows our approach to reach the second place in the SuperGLUE leaderboard\footnote{https://super.gluebenchmark.com/leaderboard} \cite{Wang2019SuperGLUEAS} between RoBERTa$_{LARGE}$ \cite{Liu2019RoBERTaAR} and the T5 model composed of 11 billions of parameters \cite{raffel2019exploring} (respectively 90.6 and 94.8 \% accuracy on the Test set).

We also notice that our method provides a much more stable training relative to the random seed as shown by the box plots in Figure~\ref{fig:copa_finetuning} a) and \ref{fig:cmqa_finetuning} a). When training on the full COPA dataset, our method exhibits a $\times 10$ standard deviation reduction on the test accuracy compared to the \textit{head CE} setting (1.35\% versus 12.8\%). Our intuition is that the improved stability is due to the better reuse of the pre-trained model priors and the absence of new randomly initialized weights. This is important result towards easier experiment comparisons as fine-tuning BERT-like architectures is known to be unstable across random restarts as shown in \cite{phang2018sentence}.

\section{Conclusions}\label{sec:concl}
In this work, we presented a new method for plausibility ranking tasks, specifically targeting commonsense ranking problem. We define a scoring function that leverages the MLM head of large pre-trained bidirectional transformer models. We establish strong results in a zero-shot setting on four commonsense reasoning datasets, comparable to supervised approaches. We then fine-tune such model using a margin-based loss on the proposed scoring function, and provide a comparative study with state of the art randomly initialized head methods. Our study demonstrates that the direct use of MLM over custom head yields increasingly superior performance gain when decreasing training data size. The proposed approach outperforms state-of-the-art training methods in terms of both test accuracy and training stability.

Future works include applying such scoring method on broader classification tasks like Natural Language Inference and Sentiment Analysis. We also think that our token-level scoring method could be used during the self-supervised pre-training phase to extend traditional next sentence prediction and sequence ordering tasks, bringing more commonsense knowledge in the model.



\begin{thebibliography}{19}
\expandafter\ifx\csname natexlab\endcsname\relax\def\natexlab#1{#1}\fi

\bibitem[{Devlin et~al.(2019)Devlin, Chang, Lee, and
  Toutanova}]{devlin2018bert}
Jacob Devlin, Ming-Wei Chang, Kenton Lee, and Kristina Toutanova. 2019.
\newblock \href {https://doi.org/10.18653/v1/N19-1423} {{BERT}: Pre-training of
  deep bidirectional transformers for language understanding}.
\newblock In \emph{Proceedings of the 2019 Conference of the North {A}merican
  Chapter of the Association for Computational Linguistics: Human Language
  Technologies, Volume 1 (Long and Short Papers)}, pages 4171--4186,
  Minneapolis, Minnesota. Association for Computational Linguistics.

\bibitem[{Gordon et~al.(2012)Gordon, Kozareva, and
  Roemmele}]{Gordon2011SemEval2012T7}
Andrew Gordon, Zornitsa Kozareva, and Melissa Roemmele. 2012.
\newblock \href {https://www.aclweb.org/anthology/S12-1052} {{S}em{E}val-2012
  task 7: Choice of plausible alternatives: An evaluation of commonsense causal
  reasoning}.
\newblock In \emph{*{SEM} 2012: The First Joint Conference on Lexical and
  Computational Semantics {--} Volume 1: Proceedings of the main conference and
  the shared task, and Volume 2: Proceedings of the Sixth International
  Workshop on Semantic Evaluation ({S}em{E}val 2012)}, pages 394--398,
  Montr{\'e}al, Canada. Association for Computational Linguistics.

\bibitem[{Holtzman et~al.(2019)Holtzman, Buys, Forbes, and
  Choi}]{Holtzman2019TheCC}
Ari Holtzman, Jan Buys, Maxwell Forbes, and Yejin Choi. 2019.
\newblock \href {https://arxiv.org/abs/1904.09751} {The curious case of neural
  text degeneration}.
\newblock \emph{ArXiv}, abs/1904.09751.

\bibitem[{Kocijan et~al.(2019)Kocijan, Cretu, Camburu, Yordanov, and
  Lukasiewicz}]{Kocijan2019ASR}
Vid Kocijan, Ana-Maria Cretu, Oana-Maria Camburu, Yordan Yordanov, and Thomas
  Lukasiewicz. 2019.
\newblock \href {https://doi.org/10.18653/v1/P19-1478} {A surprisingly robust
  trick for the {W}inograd schema challenge}.
\newblock In \emph{Proceedings of the 57th Annual Meeting of the Association
  for Computational Linguistics}, pages 4837--4842, Florence, Italy.
  Association for Computational Linguistics.

\bibitem[{Levesque et~al.(2012)Levesque, Davis, and
  Morgenstern}]{Levesque2011TheWS}
Hector~J. Levesque, Ernest Davis, and Leora Morgenstern. 2012.
\newblock \href {http://dl.acm.org/citation.cfm?id=3031843.3031909} {The
  winograd schema challenge}.
\newblock In \emph{Proceedings of the Thirteenth International Conference on
  Principles of Knowledge Representation and Reasoning}, KR'12, pages 552--561.
  AAAI Press.

\bibitem[{Li et~al.(2019)Li, Chen, and Van~Durme}]{li2019learning}
Zhongyang Li, Tongfei Chen, and Benjamin Van~Durme. 2019.
\newblock \href {https://doi.org/10.18653/v1/P19-1475} {Learning to rank for
  plausible plausibility}.
\newblock In \emph{Proceedings of the 57th Annual Meeting of the Association
  for Computational Linguistics}, pages 4818--4823, Florence, Italy.
  Association for Computational Linguistics.

\bibitem[{Liu et~al.(2019)Liu, Ott, Goyal, Du, Joshi, Chen, Levy, Lewis,
  Zettlemoyer, and Stoyanov}]{Liu2019RoBERTaAR}
Yinhan Liu, Myle Ott, Naman Goyal, Jingfei Du, Mandar Joshi, Danqi Chen, Omer
  Levy, Mike Lewis, Luke~S. Zettlemoyer, and Veselin Stoyanov. 2019.
\newblock \href {https://arxiv.org/abs/1907.11692} {Roberta: A robustly
  optimized bert pretraining approach}.
\newblock \emph{ArXiv}, abs/1907.11692.

\bibitem[{Niven and Kao(2019)}]{Niven2019ProbingNN}
Timothy Niven and Hung-Yu Kao. 2019.
\newblock \href {https://doi.org/10.18653/v1/P19-1459} {Probing neural network
  comprehension of natural language arguments}.
\newblock In \emph{Proceedings of the 57th Annual Meeting of the Association
  for Computational Linguistics}, pages 4658--4664, Florence, Italy.
  Association for Computational Linguistics.

\bibitem[{Petroni et~al.(2019)Petroni, Rockt{\"a}schel, Riedel, Lewis, Bakhtin,
  Wu, and Miller}]{petroni2019language}
Fabio Petroni, Tim Rockt{\"a}schel, Sebastian Riedel, Patrick Lewis, Anton
  Bakhtin, Yuxiang Wu, and Alexander Miller. 2019.
\newblock \href {https://doi.org/10.18653/v1/D19-1250} {Language models as
  knowledge bases?}
\newblock In \emph{Proceedings of the 2019 Conference on Empirical Methods in
  Natural Language Processing and the 9th International Joint Conference on
  Natural Language Processing (EMNLP-IJCNLP)}, pages 2463--2473, Hong Kong,
  China. Association for Computational Linguistics.

\bibitem[{Phang et~al.(2018)Phang, F{\'e}vry, and Bowman}]{phang2018sentence}
Jason Phang, Thibault F{\'e}vry, and Samuel~R Bowman. 2018.
\newblock \href {https://arxiv.org/abs/1811.01088} {Sentence encoders on
  stilts: Supplementary training on intermediate labeled-data tasks}.
\newblock \emph{arXiv preprint arXiv:1811.01088}.

\bibitem[{Poerner et~al.(2019)Poerner, Waltinger, and
  Sch{\"u}tze}]{poerner2019bert}
Nina Poerner, Ulli Waltinger, and Hinrich Sch{\"u}tze. 2019.
\newblock \href {https://arxiv.org/abs/1911.03681} {Bert is not a knowledge
  base (yet): Factual knowledge vs. name-based reasoning in unsupervised qa}.
\newblock \emph{arXiv preprint arXiv:1911.03681}.

\bibitem[{Raffel et~al.(2019)Raffel, Shazeer, Roberts, Lee, Narang, Matena,
  Zhou, Li, and Liu}]{raffel2019exploring}
Colin Raffel, Noam Shazeer, Adam Roberts, Katherine Lee, Sharan Narang, Michael
  Matena, Yanqi Zhou, Wei Li, and Peter~J Liu. 2019.
\newblock \href {https://arxiv.org/abs/1910.10683} {Exploring the limits of
  transfer learning with a unified text-to-text transformer}.
\newblock \emph{arXiv preprint arXiv:1910.10683}.

\bibitem[{Song et~al.(2019)Song, Tan, Qin, Lu, and Liu}]{song2019mass}
Kaitao Song, Xu~Tan, Tao Qin, Jianfeng Lu, and Tie-Yan Liu. 2019.
\newblock \href {http://proceedings.mlr.press/v97/song19d.html} {{MASS}: Masked
  sequence to sequence pre-training for language generation}.
\newblock In \emph{Proceedings of the 36th International Conference on Machine
  Learning}, volume~97 of \emph{Proceedings of Machine Learning Research},
  pages 5926--5936, Long Beach, California, USA. PMLR.

\bibitem[{Talmor et~al.(2019)Talmor, Herzig, Lourie, and
  Berant}]{talmor2018commonsenseqa}
Alon Talmor, Jonathan Herzig, Nicholas Lourie, and Jonathan Berant. 2019.
\newblock \href {https://doi.org/10.18653/v1/N19-1421} {{C}ommonsense{QA}: A
  question answering challenge targeting commonsense knowledge}.
\newblock In \emph{Proceedings of the 2019 Conference of the North {A}merican
  Chapter of the Association for Computational Linguistics: Human Language
  Technologies, Volume 1 (Long and Short Papers)}, pages 4149--4158,
  Minneapolis, Minnesota. Association for Computational Linguistics.

\bibitem[{Trinh and Le(2018)}]{Trinh2018ASM}
Trieu~H., Trinh and Quoc~V. Le. 2018.
\newblock \href {https://arxiv.org/abs/1806.02847} {A simple method for
  commonsense reasoning}.
\newblock \emph{ArXiv}, abs/1806.02847.

\bibitem[{Wang et~al.(2019)Wang, Pruksachatkun, Nangia, Singh, Michael, Hill,
  Levy, and Bowman}]{Wang2019SuperGLUEAS}
Alex Wang, Yada Pruksachatkun, Nikita Nangia, Amanpreet Singh, Julian Michael,
\newblock \href{http://papers.nips.cc/paper/8589-superglue-a-stickier-benchmark-for-general-purpose-language-understanding-systems.pdf}
 {Superglue: A stickier benchmark for general-purpose language understanding
  systems}.
\newblock In H.~Wallach, H.~Larochelle, A.~Beygelzimer, F.~d~Alche-Buc, E.~Fox, and R.~Garnett, editors, \emph{Advances in Neural
  Information Processing Systems 32}, pages 3261--3275. Curran Associates, Inc.

\bibitem[{Wolf et~al.(2019)Wolf, Debut, Sanh, Chaumond, Delangue, Moi, Cistac,
  Rault, Louf, Funtowicz, and Brew}]{Wolf2019HuggingFacesTS}
Thomas Wolf, Lysandre Debut, Victor Sanh, Julien Chaumond, Clement Delangue,
  Anthony Moi, Pierric Cistac, Tim Rault, R'emi Louf, Morgan Funtowicz, and
  Jamie Brew. 2019.
\newblock \href {https://arxiv.org/abs/1910.03771} {Huggingface's transformers:
  State-of-the-art natural language processing}.
\newblock \emph{ArXiv}, abs/1910.03771.

\bibitem[{Zellers et~al.(2018)Zellers, Bisk, Schwartz, and
  Choi}]{zellers2018swagaf}
Rowan Zellers, Yonatan Bisk, Roy Schwartz, and Yejin Choi. 2018.
\newblock \href {https://doi.org/10.18653/v1/D18-1009} {{SWAG}: A large-scale
  adversarial dataset for grounded commonsense inference}.
\newblock In \emph{Proceedings of the 2018 Conference on Empirical Methods in
  Natural Language Processing}, pages 93--104, Brussels, Belgium. Association
  for Computational Linguistics.

\bibitem[{Zellers et~al.(2019)Zellers, Holtzman, Bisk, Farhadi, and
  Choi}]{Zellers2019HellaSwagCA}
Rowan Zellers, Ari Holtzman, Yonatan Bisk, Ali Farhadi, and Yejin Choi. 2019.
\newblock \href {https://doi.org/10.18653/v1/P19-1472} {{H}ella{S}wag: Can a
  machine really finish your sentence?}
\newblock In \emph{Proceedings of the 57th Annual Meeting of the Association
  for Computational Linguistics}, pages 4791--4800, Florence, Italy.
  Association for Computational Linguistics.

\end{thebibliography}
\bibliographystyle{acl_natbib}


\end{document}